\documentclass[conf]{new-aiaa}
\usepackage[utf8]{inputenc}

\usepackage{graphicx}
\usepackage{amsmath}
\usepackage[version=4]{mhchem}
\usepackage[section]{placeins}
\usepackage{siunitx}
\usepackage{longtable,tabularx}
\usepackage{subcaption}
\usepackage{xcolor}
\usepackage{cleveref}
\usepackage{tikz}
\usepackage{algorithm}
\usepackage{algpseudocode}
\usepackage{hyperref}
\usepackage{tikz}
\usepackage{multirow}
\usetikzlibrary{decorations.pathreplacing}
\usepackage{comment}

\definecolor{hf}{HTML}{245B80}
\definecolor{mid}{HTML}{44877F}
\definecolor{lf}{HTML}{BB783B}
\definecolor{ink}{HTML}{303840}

\title{Multifidelity Operator Inference for Dynamical Systems}

\author{Vignesh Sella\footnote{Graduate Research Assistant, Oden
    Institute for Computational Engineering and Sciences, AIAA Student
  Member}, Anirban Chaudhuri\footnote{Research Scientist, Oden
    Institute for Computational Engineering and Sciences, AIAA Associate
  Fellow}, Karen Willcox \footnote{Director, Oden Institute for
    Computational Engineering and Sciences, Department of Aerospace
Engineering and Engineering Mechanics, AIAA Fellow}}\affil{University
of Texas at Austin, TX 78712, USA}

\begin{document}

\maketitle

\begin{abstract}
  This paper presents a multifidelity (MF) formulation of Operator
  Inference (OpInf) for learning non-intrusive, physics-based
  reduced-order models (ROMs) of nonlinear dynamical systems. OpInf
  learns reduced operators from projections of snapshot data by
  solving a least squares regression problem. However, generating
  high-fidelity snapshots can be computationally expensive, limiting
  parameter coverage and the temporal extent of training trajectories
  for many aerospace applications, which can lead to inaccurate or
  unstable ROM predictions. This work proposes MF-OpInf, which
  combines limited high-fidelity snapshots with more plentiful
  low-fidelity snapshots through a control-variates-inspired MF least
  squares formulation to learn a single set of reduced operators
  across multiple trajectories. Low-fidelity trajectories provide
  additional parameter coverage and extend beyond the high-fidelity
  training time horizon. The low-fidelity snapshots are also used to
  construct an improved reduced basis through weighted proper
  orthogonal decomposition. We demonstrate MF-OpInf on a nonlinear
  reaction-diffusion problem with parametrized sinusoidal forcing and
  a two-dimensional incompressible flow past a cylinder with a
  parametrized inlet pulse. In these experiments, under an equal
  training simulation budget, MF-OpInf reduces the mean relative
  prediction error at unseen parameters by up to $87\%$ for the
  reaction-diffusion problem and $28\%$ for the cylinder flow problem
  over the test time regime, which extends beyond the high-fidelity
  training time horizon. These results demonstrate that MF-OpInf can
  improve temporal extrapolation and prediction at unseen parameters
  without increasing the training simulation budget.
\end{abstract}

\section{Introduction}
High-fidelity (HF) numerical simulation of complex physical systems
governed by partial differential equations (PDEs) is indispensable in
aerospace engineering, yet the associated computational cost often
makes many-query applications, such as optimization, uncertainty
quantification, and real-time control, infeasible. Projection-based
reduced-order models (ROMs) address this challenge by providing
low-dimensional surrogate models that retain the relevant dynamics of
the system at a fraction of the computational
cost~\cite{benner2015survey,benner2017model,hesthaven2016certified}.
However, generating the HF snapshot data used to construct these ROMs
can be expensive, limiting parameter coverage and the duration of
training trajectories. In this work, we address this limitation
through a multifidelity (MF) approach that combines limited HF
snapshots with additional snapshots from lower-cost, low-fidelity
(LF) sources. The LF trajectories provide broader parameter coverage
and extend beyond the HF training horizon, with the aim of improving
predictions at unseen parameters and later times under a fixed
training simulation budget.

When the full-order model (FOM) implementation is unavailable or
otherwise inaccessible, for example in the case of commercial
software or legacy codes, non-intrusive methods enable ROM
construction from snapshot data without requiring access to the FOM
operators. This paper focuses on Operator Inference (OpInf), a
non-intrusive projection-based model reduction method that learns
reduced operators by solving a least squares regression
problem~\cite{PEHERSTORFER2016196}. When the training data do not
sufficiently characterize the dynamics of the system in the
prediction regime, the resulting OpInf ROM can produce inaccurate or
unstable state
estimates~\cite{mcquarrie2021data,guo2022bayesian,moore2026learning}.
Recent work has sought to improve predictive accuracy through joint
optimization of projection operators and reduced
dynamics~\cite{padovan2024data} and to enforce stability or energy
preservation through Lyapunov-based constraints on the reduced
operators~\cite{goyal2025guaranteed,gkimisis2025representation,errico2026gasnitrom}.
These approaches are complementary to the use of additional training
information considered here. Incorporating LF trajectories into
operator learning, however, requires accounting for discrepancies
between the HF and LF models.

MF methods combine information from HF and LF models by exploiting
their correlations, with the aim of improving accuracy under a given
computational budget~\cite{peherstorfer2018survey}. Several previous
approaches have incorporated MF information in the context of
data-driven model reduction, but not yet for OpInf. One class maps
the latent representations of HF and LF data, either by aligning
their reduced spaces~\cite{perron2021multi,decker2021nonlinear} or by
learning a surrogate that maps LF reduced states to their HF
counterparts~\cite{wang2020multi,conti2024multi}. A second class uses
a sequential correction strategy based on the Kennedy-O'Hagan
approach~\cite{kennedy2000predicting} in which LF predictions are
first obtained and then a discrepancy model is learned to approximate
the HF solution~\cite{demo2023deeponet,sella2025improving}. A third
class pools HF and LF data directly into a weighted regression
problem, where LF samples are assigned reduced influence relative to
HF observations~\cite{sella2025projection} for linear regression. In
these approaches, LF information enters either through the
construction of the reduced space, as a prior prediction to be
corrected, or as down-weighted observations in the regression.
Another way to combine HF and LF information is through control
variates, as used in MF approaches for Monte Carlo variance
reduction~\cite{ng2013multidelity,peherstorfer2016optimal,giles2008multilevel,giles2015multilevel,Gorodetsky2020ACV,schaden2020multilevel}.
The regression formulation of OpInf provides a natural setting for
this approach, allowing LF information to enter through a correction
to the HF fitting objective while preserving the form of the reduced
dynamical model.
\begin{figure}[!htb]
  \centering
  \includegraphics{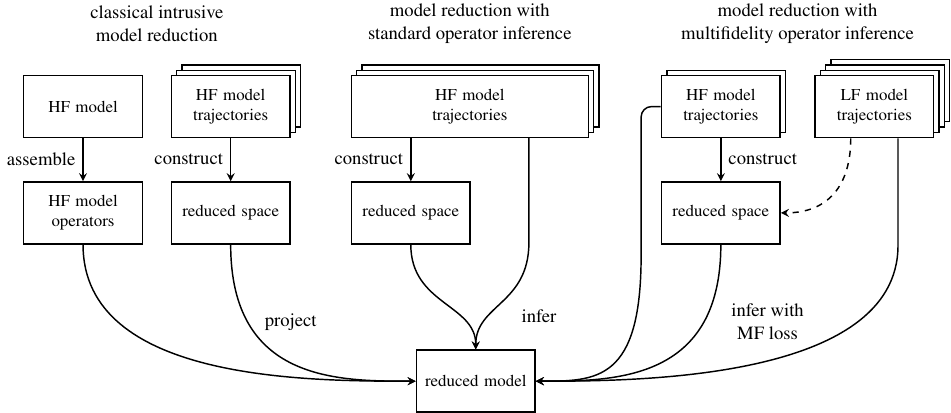}
  \caption{Illustrative diagram showcasing the difference between
    intrusive model reduction, standard OpInf, and the proposed
    MF-OpInf. The projection step in the intrusive approach requires
    the operators of the full model in order to construct the reduced
    operators, whereas OpInf and MF-OpInf only require data from the
    FOM and the form of the governing equations. In the MF approach the
    LF model trajectories are used in the learning of the reduced model
    and can also be used to construct the reduced basis. Figure adapted
  from~\cite{PEHERSTORFER2016196}.}
  \label{fig:mfopinf_overview}
\end{figure}

Here, we develop MF-OpInf for high-dimensional nonlinear dynamical
systems, preserving the structure of the governing equations through
the OpInf framework. The proposed method incorporates a control
variate correction into the OpInf least squares regression problem.
Data from all fidelity levels are represented in one reduced space
and used to infer a single set of reduced operators. We also utilize
LF snapshots to construct the reduced basis through weighted proper
orthogonal decomposition (w-POD)~\cite{christensen1999evaluation}.
This choice of basis is distinct from the control-variate correction
used in operator fitting. An overview of how our formulation compares
with classical intrusive model reduction and standard OpInf is
provided in~\Cref{fig:mfopinf_overview}. We limit the scope of this
work to consider systems with parametrized inputs or initial
conditions, for which the training data consist of multiple
trajectories but a single set of parameter-independent reduced
operators is learned for the ROM. This setting does not consider the
general parametric OpInf case
\cite{PEHERSTORFER2016196,mcquarrie2023nonintrusive}, where the
reduced operators explicitly depend on the parameters.

This work presents the first MF formulation for learning reduced
operators within the OpInf framework. The key contributions are as follows:
\begin{enumerate}
  \item We develop a control-variate-based MF least squares objective
    that combines limited HF trajectories with more plentiful LF
    trajectories to infer a single set of reduced operators. The
    formulation accommodates multiple fidelity levels and snapshot
    trajectories with different training horizons.
  \item We derive the closed-form expression for the regularized
    normal equations for MF-OpInf.
  \item We investigate budget allocations that trade either the
    number or the duration of HF trajectories for additional LF simulations.
  \item We evaluate MF-OpInf against HF-only OpInf given an equal
    training simulation budget on two nonlinear dynamical systems. We
    assess predictions at unseen parameter values and beyond the HF
    training horizon, and examine the effects of reduced dimension
    and the use of LF snapshots in w-POD~\cite{christensen1999evaluation}.
\end{enumerate}
We demonstrate the MF-OpInf method on two benchmark problems: a
nonlinear reaction-diffusion equation with parametrized sinusoidal
forcing and two-dimensional incompressible flow past a cylinder. The
reaction-diffusion problem tests prediction across forcing amplitudes
and beyond the HF training horizon, with LF training trajectories
extending through the prediction interval. The cylinder problem uses
a parametrized inlet pulse to generate different developing wakes and
tests prediction of the subsequent flow beyond both the HF and LF
training horizons. The results show the effectiveness of MF-OpInf at
utilizing additional LF training information to improve predictions
at unseen parameters and later times.

The remainder of this manuscript is organized as follows.
\Cref{sec:opinf_background} briefly establishes the relevant
background on the standard OpInf methodology. \Cref{sec:mfopinf}
presents the proposed MF-OpInf formulation, reduced basis choices,
and implementation details. \Cref{sec:experiments} describes the
numerical experiments and compares the prediction accuracy of HF-only
and MF-OpInf under a common training simulation budget.
\Cref{sec:conclusion} summarizes our contributions and discusses
directions for future work.

\section{Reduced-order models using operator inference}
\label{sec:opinf_background}
OpInf~\cite{PEHERSTORFER2016196} is a data-driven approach to
non-intrusively learn a projection-based ROM from simulation data,
wherein the structure of the ROM is enforced by the known governing
equations. To illustrate, we consider the following FOM governing
equations for a dynamical system with quadratic model form, spatially
discretized in semi-discrete form,
\begin{align}
  \label{eq:FOM}
  \frac{\text{d}}{\text{d}t}\mathbf{q}(t;p) = \mathbf{c} +
  \mathbf{A}\mathbf{q}(t;p) + \mathbf{H}\big(\mathbf{q}(t;p) \otimes
  \mathbf{q}(t;p)\big) + \mathbf{B}\mathbf{u}(t;p), \quad
  \mathbf{q}(t_0;p) = \mathbf{q}_0(p), \quad t \in [t_0, t_f],
\end{align}
where $p\in \mathcal{P}$ is a trajectory-generating parameter that
specifies the initial condition, the input, or both, $\mathbf{q}(t;p)
\in \mathbb{R}^n$ is the $n$-dimensional spatially discretized state
at time $t$, $\mathbf{u}(t;p) \in \mathbb{R}^m$ are the $m$ inputs at
time $t$ (such as time-dependent source terms and/or boundary
conditions), $t_0$ and $t_f$ are the initial and final time, and
$\mathbf{q}_0(p) \in \mathbb{R}^n$ is the initial condition. In
Eq.~\eqref{eq:FOM}, $\mathbf{c} \in \mathbb{R}^n$ is a constant
vector, $\mathbf{A q}(t;p)$ is linear in the state $\mathbf{q}(t;p)$
with $\mathbf{A} \in \mathbb{R}^{n \times n}$, $
\mathbf{H}(\mathbf{q}(t;p) \otimes \mathbf{q}(t;p))$ are terms
quadratic in the state with $\mathbf{H} \in \mathbb{R}^{n \times
n^2}$, and $\mathbf{B u}(t;p)$ are linear in the input
$\mathbf{u}(t;p)$ with $\mathbf{B} \in \mathbb{R}^{n \times m}$.
Note, the operators $\mathbf{c},\mathbf{A},\mathbf{H},\mathbf{B}$ do
not depend on $p$, hence, a single set of reduced operators can be
learned from trajectories generated with different values of $p$. In
addition, we are using a quadratic model form for ease of exposition;
however, OpInf can be applied to any model form that leads to a
linear least squares problem as discussed below.

For the time-dependent problems of interest, the FOM state is often
high-dimensional, making repeated direct simulation expensive for
many-query problems. Thus, a projection-based ROM, which preserves
the structure of the FOM, is employed. We collect training data from
$s$ trajectories with parameter values $P=\{p_1,\dots,p_s\} \subset
\mathcal{P}$ each sampled at the same $K$ times
$t_0<t_1<\cdots<t_{K-1}$ in $[t_0,t_f]$. Let
$\mathbf{q}_{p_j,i}:=\mathbf{q}(t_i;p_j)$ denote the snapshot of
trajectory $p_j$ at time $t_i$. The trajectory and training snapshot
matrices are
\begin{align*}
  \mathbf{Q}_{p_j}=
  \begin{bmatrix}\mathbf{q}_{p_j,0}&\cdots&\mathbf{q}_{p_j,K-1}
  \end{bmatrix}\in\mathbb{R}^{n\times K},
  \qquad
  \mathbf{Q}_{P}=
  \begin{bmatrix}\mathbf{Q}_{p_1}&\cdots&\mathbf{Q}_{p_s}
  \end{bmatrix}\in\mathbb{R}^{n\times sK}.
\end{align*}
In standard OpInf, a reduced basis is typically constructed using POD
of $\mathbf{Q}_P$. The thin singular value decomposition (SVD) of
this matrix gives
\begin{align}
  \mathbf{Q}_P=\boldsymbol{\Phi}\boldsymbol{\Sigma}\boldsymbol{\Psi}^{\top},
  \qquad \mathbf{V}=[\boldsymbol{\phi}_1\;\cdots\;\boldsymbol{\phi}_k],
  \qquad \mathbf{V}^{\top}\mathbf{V}=\mathbf{I}_k,
  \label{eq:pod_svd}
\end{align}
where the columns of $\boldsymbol{\Phi}$ and $\boldsymbol{\Psi}$ are
the left and right singular vectors respectively, and
$\boldsymbol{\Sigma}$ is a diagonal matrix containing the singular
values. The leading $k$ left singular vectors form the POD basis
$\mathbf{V}\in\mathbb{R}^{n\times k}$, which minimizes
$\|\mathbf{Q}_P-\mathbf{V}\mathbf{V}^{\top}\mathbf{Q}_P\|_F^2$ over
all bases with $k$ orthonormal columns. If centering is used, the
mean over all $sK$ snapshots is subtracted before computing the SVD.
In the uncentered formulation, the state is approximated as
$\mathbf{q}(t;p)\approx\mathbf{V}\mathbf{z}(t;p)$, where
$\mathbf{z}(t;p)=\mathbf{V}^{\top}\mathbf{q}(t;p)\in\mathbb{R}^k$ and
$k\ll n$. The ROM has the same form as the FOM in Eq.~\eqref{eq:FOM},
\begin{align*}
  \frac{\text{d}}{\text{d}t}
  \mathbf{z}(t;p)=\widehat{\mathbf{c}}+\widehat{\mathbf{A}}
  \mathbf{z}(t;p)+\widehat{\mathbf{H}}\big(\mathbf{z}(t;p) \otimes
  \mathbf{z}(t;p)\big)+\widehat{\mathbf{B}} \mathbf{u}(t;p), \quad
  \mathbf{z}(t_0;p)=\mathbf{V}^{\top} \mathbf{q}_0(p), \quad t \in[t_0, t_f],
\end{align*}
where $\widehat{\mathbf{c}} \in \mathbb{R}^k, \widehat{\mathbf{A}}
\in \mathbb{R}^{k \times k}, \widehat{\mathbf{H}} \in \mathbb{R}^{k
\times\binom{k+1}{2}}$, and $\widehat{\mathbf{B}} \in \mathbb{R}^{k
\times m}$ are determined by solving a data-driven regression
problem. In practice, to remove redundant terms in the reduced
representation of the quadratic operator $\widehat{\mathbf{H}}$, we
employ the compact Kronecker product that leads to a matrix size of
$k \times\binom{k+1}{2}$.

Let $\mathbf{z}_{p_j,i}=\mathbf{V}^\top\mathbf{q}_{p_j,i}$ denote the
reduced snapshots, $\dot{\mathbf{z}}_{p_j,i}$ the corresponding time
derivatives, and $\mathbf{u}_{p_j,i}:=\mathbf{u}(t_i;p_j)$ the
inputs. The time derivatives are estimated separately for each
trajectory, e.g., by finite differences within each trajectory. The
operators are learned by solving the regularized least-squares problem,
\begin{align}
  \label{eq:opinf_sum}
  \min_{\widehat{\mathbf{c}}, \widehat{\mathbf{A}},
  \widehat{\mathbf{H}}, \widehat{\mathbf{B}}}
  \frac{1}{s}\sum_{j=1}^{s}\frac{1}{K}\sum_{i=0}^{K-1}\left\|\widehat{\mathbf{c}}+\widehat{\mathbf{A}}
  \mathbf{z}_{p_j,i}+\widehat{\mathbf{H}}\left(\mathbf{z}_{p_j,i}
  \otimes\mathbf{z}_{p_j,i}\right)+\widehat{\mathbf{B}}
  \mathbf{u}_{p_j,i}-\dot{\mathbf{z}}_{p_j,i}\right\|_2^2 +
  \left\|\mathbf{\Gamma O}^{\top}\right\|_F^2,
\end{align}
where $\mathbf{O}$ collects the unknown operators (defined below). In
essence, the OpInf regression problem finds the reduced operators
that best match the projected data in the minimum $\ell_2$-norm
sense, averaged over snapshots and trajectories. To write
Eq.~\eqref{eq:opinf_sum} in matrix form~\cite{mcquarrie2021data},
define for each trajectory $p_j$
\begin{align}
  & \mathbf{D}_{p_j}=
  \begin{bmatrix}\mathbf{1}_{K} & \mathbf{Z}_{p_j}^{\top} &
    (\mathbf{Z}_{p_j} \otimes \mathbf{Z}_{p_j})^{\top} & \mathbf{U}_{p_j}^{\top}
  \end{bmatrix} \in \mathbb{R}^{K \times d(k, m)}, && \text{(known
  data)} \label{eq:known_data} \\
  & \mathbf{Z}_{p_j}=
  \begin{bmatrix}\mathbf{z}_{p_j,0} & \cdots & \mathbf{z}_{p_j,K-1}
  \end{bmatrix} \in \mathbb{R}^{k \times K}, &&\text{(reduced
  snapshots)} \label{eq:reduced_snapshots} \\
  & \mathbf{R}_{p_j}=
  \begin{bmatrix}\dot{\mathbf{z}}_{p_j,0} & \cdots & \dot{\mathbf{z}}_{p_j,K-1}
  \end{bmatrix} \in \mathbb{R}^{k \times K}, &&\text{(time
  derivatives)} \label{eq:time_derivatives} \\
  & \mathbf{U}_{p_j}=
  \begin{bmatrix}\mathbf{u}_{p_j,0} & \cdots & \mathbf{u}_{p_j,K-1}
  \end{bmatrix} \in \mathbb{R}^{m \times K}, &&\text{(inputs)} \label{eq:inputs}
\end{align}
and the unknown operators $\mathbf{O}=
\begin{bmatrix}\widehat{\mathbf{c}} & \widehat{\mathbf{A}} &
  \widehat{\mathbf{H}} & \widehat{\mathbf{B}}
\end{bmatrix} \in \mathbb{R}^{k \times d(k, m)}$. Here $d(k,m) = 1 +
k + \binom{k+1}{2} + m$ and $\mathbf{1}_{K} \in \mathbb{R}^K$ is a
column vector with all entries set to unity. Additionally, define the
concatenated data
\begin{align}
  \label{eq:D_and_R}
  \mathbf{D}_{P}=
  \begin{bmatrix}
    \mathbf{D}_{p_1}\\ \vdots\\ \mathbf{D}_{p_{s}}
  \end{bmatrix}
  \in\mathbb{R}^{sK\times d(k,m)}, \qquad
  \mathbf{R}_{P}=
  \begin{bmatrix}\mathbf{R}_{p_1}&\cdots&\mathbf{R}_{p_{s}}
  \end{bmatrix}\in\mathbb{R}^{k\times sK}.
\end{align}
Equation~\eqref{eq:opinf_sum} in matrix form is then,
\begin{align}
  \label{eq:regularized_opinf}
  \min _{\mathbf{O}}\; \widehat{\mathcal{L}}_{P}(\mathbf{O}) +
  \left\|\mathbf{\Gamma O}^{\top}\right\|_F^2,
  \qquad
  \widehat{\mathcal{L}}_{P}(\mathbf{O}) :=
  \frac{1}{sK}\left\|\mathbf{D}_{P}\mathbf{O}^{\top}-\mathbf{R}_{P}^{\top}\right\|_F^2,
\end{align}
where $\mathbf{\Gamma} \in \mathbb{R}^{d(k,m)\times d(k,m)}$ is the
full-rank regularizer. The regularizer is a function of multiple
regularization parameters, which are generally chosen to penalize the
operators ($\widehat{\mathbf{c}}, \widehat{\mathbf{A}}$),
$\widehat{\mathbf{B}}$, and $\widehat{\mathbf{H}}$, respectively. The
regularization is introduced to help with an ill-conditioned data
matrix and to avoid over-fitting.

\section{MF-OpInf: A multifidelity approach for operator inference}
\label{sec:mfopinf}
We now present our MF formulation for OpInf that incorporates MF
snapshot training data into the OpInf framework. We begin with the
multifidelity sampling setup for the MF least squares formulation
considered in this work, and then provide a description of the
MF-OpInf approach. We then discuss the reduced basis construction
using MF data followed by implementation details.

\subsection{Multifidelity setup for a dynamical system}
\label{sec:MFMC_setup}
In this work, we treat a complete trajectory, obtained by evaluating
one of the available data-generating models at a specified parameter
value, as one sample. Let $L$ denote the number of levels of
fidelity, and let $\mathbf{S}=[s_1,\dots,s_L]^\top\in\mathbb{N}^L$,
where $s_l$ is the number of trajectories evaluated at levels
$l=1,\dots,L$ and $0<s_1\leq\cdots\leq s_L$. The HF model is always
denoted with $l=1$, and each model after is lower-fidelity. For each
level $l$, let $P_l=\{p_1,\dots,p_{s_l}\}$ denote the corresponding
set of trajectory-generating parameter values, with
$P_1\subseteq\cdots\subseteq P_L$. Each $p_j$ may specify an initial
condition, an input parameter, or a combination thereof. We visually
depict the nested sampling in~\Cref{fig:nested_parameter_sets}. This
nested sampling strategy is inspired by the control-variates-based
multifidelity Monte Carlo (MFMC) method, which is used to reduce
Monte Carlo error for a fixed computational budget by combining
information from models of varying
fidelity~\cite{ng2013multidelity,peherstorfer2016optimal,hyun2026robust}.
We use the control variates method to formulate the MF-OpInf least
squares problem, in which samples from an HF source and one or more
LF sources are incorporated directly into the operator learning
problem~\eqref{eq:regularized_opinf}.

In this paper, we consider that every trajectory evaluated at level
$l$ contains the same number of snapshots, $K_l$. Thus, $s_lK_l$ is
the total number of snapshots available at that level. The snapshot
times and the number of snapshots per trajectory may differ across
levels. The nested sampling assumption applies only to the sets of
trajectory-generating parameter values and does not require nested
sets of snapshot times. For each shared parameter value $p_j$,
trajectories evaluated at different levels must correspond to the
same parameter, initial condition, or input function. Note that,
whereas our approach is built upon the MFMC setup, the addition of LF
data to the least squares objective is compatible with other MF
control variate methods such as multilevel Monte Carlo
(MLMC)~\cite{giles2008multilevel,giles2015multilevel}, approximate
control variates~\cite{Gorodetsky2020ACV}, and multilevel best linear
unbiased estimators
(MLBLUE)~\cite{schaden2020multilevel,schaden2021asymptotic,croci2023multi}.
\begin{figure}[!htb]
  \centering
  \begin{tikzpicture}[x=1cm,y=1cm,scale=.92,transform shape,
    every node/.style={font=\small,text=ink},
    dot/.style={circle,inner sep=0pt,minimum size=4.8pt},
  region/.style={line width=.8pt}]
  \path[use as bounding box] (-.15,-.15) rectangle (6.25,2.98);

  \draw[draw=ink!65,line width=1.0pt,rounded corners=0pt]
  (-.12,-.12) rectangle (6.22,2.95);

  \draw[region,draw=lf,fill=lf!7]
  (.15,1.25) .. controls (-.05,2.15) and (1.05,2.55) .. (2.2,2.35)
  .. controls (3.25,2.12) and (4.2,2.65) .. (5.35,2.27)
  .. controls (6.25,1.98) and (6.3,.6) .. (5.46,.25)
  .. controls (4.55,-.12) and (3.6,.3) .. (2.65,.16)
  .. controls (1.42,-.05) and (.25,.23) .. (.15,1.25) -- cycle;

  \draw[region,draw=mid,fill=mid!9]
  (.47,1.25) .. controls (.35,1.95) and (1.03,2.18) .. (1.85,2.06)
  .. controls (2.62,1.91) and (3.22,2.22) .. (3.91,1.8)
  .. controls (4.5,1.42) and (4.25,.65) .. (3.6,.5)
  .. controls (2.88,.31) and (2.43,.62) .. (1.71,.46)
  .. controls (.88,.3) and (.55,.63) .. (.47,1.25) -- cycle;

  \draw[region,draw=hf,fill=hf!11]
  (.73,1.19) .. controls (.64,1.69) and (1.12,1.91) .. (1.66,1.76)
  .. controls (2.13,1.63) and (2.4,1.68) .. (2.5,1.2)
  .. controls (2.63,.71) and (2.05,.65) .. (1.61,.76)
  .. controls (1.08,.88) and (.87,.71) .. (.73,1.19) -- cycle;

  \node[text=hf] at (1.51,1.57) {$P_1$};
  \node[text=mid] at (3.25,1.74) {$P_2$};
  \node[text=lf] at (5.17,2.05) {$P_3$};

  \foreach \x/\y/\j in {
    1.13/1.25/1,
    2.04/1.23/2,
    2.94/1.25/3,
    3.65/.95/4,
    3.80/1.55/5,
    4.80/1.50/6,
    5.42/.82/7,
    4.65/.65/8,
    5.55/1.65/9,
    4.15/2.15/10
  } {
    \node[dot,fill=ink,minimum size=3.6pt] at (\x,\y) {};
    \node[anchor=north,font=\footnotesize,inner sep=2pt]
    at (\x,\y-.04) {$p_{\j}$};
  }

  \node[anchor=north west,font=\normalsize,inner sep=0pt]
  at (.08,2.80) {Parameter space};

  \node[overlay,anchor=west,align=left,inner sep=0pt, font=\large]
  at (6.55,1.42) {%
    $P_l=\{p_i\}_{i=1}^{s_l},\quad l \in \{1,2,3\}$\\[3pt]
    $s_1=2,\quad s_2=5,\quad s_3=10$
  };
\end{tikzpicture}
  \caption{Illustrative diagram of the nested sampling structure used
    in the MF-OpInf setup. Parameters $p_i$ refer to values used within
    the source terms, boundary conditions, or initial conditions of the
    dynamical system. The sets of parameters satisfy $P_1\subseteq
    P_2\subseteq P_3$, while the contours indicate the values available
  at each fidelity level.}
  \label{fig:nested_parameter_sets}
\end{figure}

\subsection{Control-variate-based loss function formulation for MF-OpInf}
\label{sec:mf_opinf_bg}

For each level $l$, we construct the matrices introduced in
Eqs.~\eqref{eq:known_data}-\eqref{eq:inputs} from the level-$l$ data,
with $(s,K,P)$ replaced by $(s_l,K_l,P_l)$ and a superscript $(l)$
added to reflect which model the snapshots were obtained from. For
each trajectory, we define
$\mathbf{D}^{(l)}_{p_j}\in\mathbb{R}^{K_l\times d(k,m)}$,
$\mathbf{Z}^{(l)}_{p_j}\in\mathbb{R}^{k\times K_l}$,
$\mathbf{R}^{(l)}_{p_j}\in\mathbb{R}^{k\times K_l}$, and
$\mathbf{U}^{(l)}_{p_j}\in\mathbb{R}^{m\times K_l}$. For the $s_l$
trajectories with parameters in $P_l$, we define the concatenated
matrices $\mathbf{D}^{(l)}_{P_l}\in\mathbb{R}^{s_lK_l\times d(k,m)}$
and $\mathbf{R}^{(l)}_{P_l}\in\mathbb{R}^{k\times s_lK_l}$ as in
Eq.~\eqref{eq:D_and_R}.
For $l\geq 2$, we also use the level-$l$ data restricted to the
$s_{l-1}$ trajectories with parameters in $P_{l-1}\subseteq P_l$,
denoted $\mathbf{D}^{(l)}_{P_{l-1}}\in\mathbb{R}^{s_{l-1}K_l\times
d(k,m)}$ and $\mathbf{R}^{(l)}_{P_{l-1}}\in\mathbb{R}^{k\times s_{l-1}K_l}$.
The unknown reduced operators $\mathbf{O}\in\mathbb{R}^{k\times
d(k,m)}$ remain as defined in Sec.~\ref{sec:opinf_background}. The
regression problem for MF-OpInf seeks reduced operators that minimize
the loss function defined over these concatenated matrices across
different fidelity levels. The fidelity-specific loss averages the
squared residual over snapshots and trajectories:
\begin{align}
  \widehat{\mathcal{L}}_{P_l}^{(l)}(\mathbf{O})&:=\frac{1}{s_l}\sum_{j=1}^{s_l}\frac{1}{K_l}\left\|\mathbf{D}_{p_j}^{(l)}\mathbf{O}^\top-\mathbf{R}_{p_j}^{(l)\top}\right\|_F^2
  =\frac{1}{s_lK_l}\left\|\mathbf{D}_{P_l}^{(l)}\mathbf{O}^\top-\mathbf{R}_{P_l}^{(l)\top}\right\|_F^2,
  && l=1,\dots,L, \label{eq:fidelity_loss}\\
  \widehat{\mathcal{L}}_{P_{l-1}}^{(l)}(\mathbf{O})&:=\frac{1}{s_{l-1}K_l}\left\|\mathbf{D}_{P_{l-1}}^{(l)}\mathbf{O}^\top-\mathbf{R}_{P_{l-1}}^{(l)\top}\right\|_F^2,
  && l=2,\dots,L. \nonumber
\end{align}
The normalization in the above loss gives equal weight to all
snapshots within a fidelity level and, since each trajectory at
fidelity $l$ contains $K_l$ snapshots, equal weight to each
trajectory. This average loss is equivalent to concatenating the
trajectories and averaging over all $s_lK_l$ snapshots. We choose
this approach for ease of exposition. Alternatively, within the same
fidelity level, data corresponding to different trajectories could
have a different number of snapshots, but additional indexing would be required.

We develop a regularized MF loss that combines the fidelity-specific
losses defined by Eq.~\eqref{eq:fidelity_loss} to utilize information
from multiple models as
\begin{equation}
  \label{eq:MF_loss_multi}
  \widehat{\mathcal{L}}_{\text{MF}}(\mathbf{O},\boldsymbol{\alpha})
  =\widehat{\mathcal{L}}_{P_1}^{(1)}(\mathbf{O})
  +\sum_{l=2}^{L}\alpha_l\left[\widehat{\mathcal{L}}_{P_l}^{(l)}(\mathbf{O})-\widehat{\mathcal{L}}_{P_{l-1}}^{(l)}(\mathbf{O})\right]
  +\left\|\mathbf{\Gamma}\mathbf{O}^\top\right\|_F^2,
\end{equation}
where each component of
$\boldsymbol{\alpha}=(\alpha_2,\dots,\alpha_L)$ is a scalar control
variate coefficient. The first term in Eq.~\eqref{eq:MF_loss_multi}
provides the HF data regression baseline. In each subsequent
correction term, both losses are evaluated at fidelity $l$, using the
parameter sets $P_l$ and $P_{l-1}$, respectively. Each correction
compares the average fitting error over the larger LF parameter set
with that over the nested subset, using LF data to adjust the HF
fitting objective.
The coefficients $\boldsymbol{\alpha}$ control the influence of LF
snapshots on the inferred reduced operators. Rather than estimating
them from model correlations using pilot samples, as in
MFMC~\cite{peherstorfer2016optimal}, we treat $\boldsymbol{\alpha}$
as a tunable hyperparameter selected by the validation procedure
in~\Cref{sec:implementation}. A systematic study of
$\boldsymbol{\alpha}$ and a theoretical analysis of its effect are
left for future work.
For $L=1$, Eq.~\eqref{eq:MF_loss_multi} reduces to the
single-fidelity problem~\eqref{eq:regularized_opinf} with $P=P_1$.
Our prior work~\cite{qian2024multifidelity} applied control variates
to the cross-covariance term of the loss function to learn linear
regression models from MF data for a scalar output. Here, we apply a
control-variate-inspired formulation to the entire OpInf loss
function to learn reduced operators for high-dimensional dynamical systems.

To derive the normal equations, note that each fidelity-specific loss
has the same form as seen in Eq.~\eqref{eq:fidelity_loss}. Its
gradient with respect to the reduced operators is given by
\begin{equation*}
  \frac{\partial}{\partial\mathbf{O}^\top}\widehat{\mathcal{L}}_{P_l}^{(l)}=\frac{2}{s_lK_l}\left(\mathbf{D}_{P_l}^{(l)}\right)^\top\left(\mathbf{D}_{P_l}^{(l)}\mathbf{O}^\top-\mathbf{R}_{P_l}^{(l)\top}\right),
  \qquad
  \frac{\partial}{\partial\mathbf{O}^\top}\left\|\mathbf{\Gamma}\mathbf{O}^\top\right\|_F^2=2\mathbf{\Gamma}^\top\mathbf{\Gamma}\mathbf{O}^\top,
\end{equation*}
and the gradient of $\widehat{\mathcal{L}}_{P_{l-1}}^{(l)}$ follows
by replacing $P_l$ and $s_l$ with $P_{l-1}$ and $s_{l-1}$. Thus,
differentiating Eq.~\eqref{eq:MF_loss_multi} yields
\begin{align*}
  \frac{1}{2}\frac{\partial\widehat{\mathcal{L}}_{\text{MF}}}{\partial\mathbf{O}^\top}
  &=\frac{1}{s_1K_1}\left(\mathbf{D}_{P_1}^{(1)}\right)^\top
  \left(\mathbf{D}_{P_1}^{(1)}\mathbf{O}^\top-\mathbf{R}_{P_1}^{(1)\top}\right)\\
  &\quad+\sum_{l=2}^{L}\alpha_l\left[
    \frac{1}{s_lK_l}\left(\mathbf{D}_{P_l}^{(l)}\right)^\top
    \left(\mathbf{D}_{P_l}^{(l)}\mathbf{O}^\top-\mathbf{R}_{P_l}^{(l)\top}\right)
    -\frac{1}{s_{l-1}K_l}\left(\mathbf{D}_{P_{l-1}}^{(l)}\right)^\top
  \left(\mathbf{D}_{P_{l-1}}^{(l)}\mathbf{O}^\top-\mathbf{R}_{P_{l-1}}^{(l)\top}\right)\right]\\
  &\quad+\mathbf{\Gamma}^\top\mathbf{\Gamma}\mathbf{O}^\top.
\end{align*}
Setting the derivative to zero and collecting terms leads to the
following normal equations for MF-OpInf:
\begin{equation}
  \left(\mathbf{G}_{\text{MF}}+\mathbf{\Gamma}^\top\mathbf{\Gamma}\right)\mathbf{O}^\top=\mathbf{F}_{\text{MF}},
  \label{eq:mf_opinf_multi_normal_equations}
\end{equation}
where
\begin{align}
  \mathbf{G}_{\text{MF}}
  &:=\frac{1}{s_1K_1}\left(\mathbf{D}_{P_1}^{(1)}\right)^\top\mathbf{D}_{P_1}^{(1)}
  +\sum_{l=2}^{L}\alpha_l\left[
    \frac{1}{s_lK_l}\left(\mathbf{D}_{P_l}^{(l)}\right)^\top\mathbf{D}_{P_l}^{(l)}
  -\frac{1}{s_{l-1}K_l}\left(\mathbf{D}_{P_{l-1}}^{(l)}\right)^\top\mathbf{D}_{P_{l-1}}^{(l)}\right],
  \label{eq:mf_opinf_multi_A}\\
  \mathbf{F}_{\text{MF}}
  &:=\frac{1}{s_1K_1}\left(\mathbf{D}_{P_1}^{(1)}\right)^\top\mathbf{R}_{P_1}^{(1)\top}
  +\sum_{l=2}^{L}\alpha_l\left[
    \frac{1}{s_lK_l}\left(\mathbf{D}_{P_l}^{(l)}\right)^\top\mathbf{R}_{P_l}^{(l)\top}
  -\frac{1}{s_{l-1}K_l}\left(\mathbf{D}_{P_{l-1}}^{(l)}\right)^\top\mathbf{R}_{P_{l-1}}^{(l)\top}\right].
  \label{eq:mf_opinf_multi_B}
\end{align}
Note that $\mathbf{G}_{\text{MF}}$ is a control-variates-inspired
combination of fidelity-specific Gram matrices, while
$\mathbf{F}_{\text{MF}}$ is the corresponding combination of the
cross-products of the data and time-derivatives
$\mathbf{D^\top\mathbf{R}^\top}$. The reduced operators are obtained
by solving the normal equations in
Eq.~\eqref{eq:mf_opinf_multi_normal_equations}. In the MF setting,
when $L\geq 2$, $\mathbf{G}_{\text{MF}}$ alone need not be positive
semidefinite due to the control-variate correction terms. Hence, we
require $\mathbf{G}_{\text{MF}}+\mathbf{\Gamma}^\top\mathbf{\Gamma}$
to be positive definite, in which case the regularized objective has
the unique minimizer given by
\begin{equation}
  \label{eq:mfopinf_operator}
  \mathbf{O}^\top=\left(\mathbf{G}_{\text{MF}}+\mathbf{\Gamma}^\top\mathbf{\Gamma}\right)^{-1}\mathbf{F}_{\text{MF}}.
\end{equation}

\subsection{Reduced basis selection with multifidelity data} \label{sec:reduced}
Our MF-OpInf formulation requires data from all fidelity levels to be
represented in a common reduced space. We consider two choices: a POD
basis constructed from HF snapshots only, and a w-POD basis that also
includes LF snapshots. The choice of dimensionality reduction is
problem dependent, and the effectiveness of POD depends on whether a
sufficiently low-dimensional linear (or affine) representation of the
states is adequate. Since LF trajectories are already available for
MF operator inference, we explore their use in basis construction
without requiring additional FOM simulations. In particular, LF
trajectories at additional parameter values and over longer training
horizons can contain information about states not represented in the
HF training snapshots. Including this information in the basis
construction can reduce the projection error over the prediction
horizon. For a fixed approximation space, operator inference cannot
recover state components outside that space, so this representation
error limits the accuracy of the ROM regardless of the inferred operators.

At a fixed basis dimension, including additional states can improve
representation beyond the HF training regime at the expense of
accuracy on the HF training data. We note that LF model error may
further affect the quality of the resulting basis. We therefore
assess basis construction empirically as an implementation choice
alongside MF operator fitting. w-POD is one possible choice for
incorporating LF data. Note that the MF-OpInf approach is also
amenable to other ways of constructing the basis from MF data.

\paragraph{HF POD basis.}
The first choice of basis we employ is the HF POD basis, obtained
from the HF training snapshots only. With this choice, LF data are
used only for operator fitting, discarding the LF information for
basis construction. Under the same training simulation budget, the MF
method uses fewer HF training snapshots to accommodate the cost of LF
data generation. Consequently, the HF POD basis used by the MF
strategy may differ from the HF-only strategy because they are
constructed from different HF training datasets. We found that the HF
POD strategy does not perform as well when used with MF-OpInf since
it does not use LF information in basis construction (see
\Cref{sec:basis_comp}).

\paragraph{Weighted POD basis.}
The second choice is to use w-POD, which allows us to include LF
snapshots in the basis construction. Christensen et
al.~\cite{christensen1999evaluation} considered w-POD to give greater
priority to selected features of parameter-dependent flows. In this
work, we adapt w-POD for MF data and assign weights according to
snapshot fidelity. Here, we describe the construction for the
bi-fidelity case. Let $\mathbf{Q}_{\text{HF}}$ and
$\mathbf{Q}_{\text{LF}}$ contain the HF and LF training snapshots,
respectively, with the LF snapshots interpolated onto the HF spatial
grid. We assign weight $1/(s_1K_1)$ to each HF snapshot and weight
$w/(s_2K_2)$ to each LF snapshot, where $w\geq0$. The factors
$1/(s_1K_1)$ and $1/(s_2K_2)$ account for the total number of
snapshots at each fidelity level, so that the relative contribution
of LF data is controlled by $w$ even when more LF trajectories or
snapshots per trajectory are available. If centering is used, both
matrices contain snapshots centered about the same weighted mean
$\overline{\mathbf{q}}_w=(\overline{\mathbf{q}}_{\text{HF}}+w\overline{\mathbf{q}}_{\text{LF}})/(1+w)$,
where $\overline{\mathbf{q}}_{\text{HF}}$ and
$\overline{\mathbf{q}}_{\text{LF}}$ are the means of the snapshots at
each fidelity. Following the snapshot-weighted POD formulation
in~\cite{grassle2018pod}, define
\begin{equation}
  \mathbf{Q}_w
  =
  \left[\frac{1}{\sqrt{s_1K_1}}\mathbf{Q}_{\text{HF}}\;\middle|\;
  \sqrt{\frac{w}{s_2K_2}}\mathbf{Q}_{\text{LF}}\right].
  \label{eq:weighted_snapshot_matrix}
\end{equation}
For any $\mathbf{V}\in\mathbb{R}^{n\times k}$ with
$\mathbf{V}^{\top}\mathbf{V}=\mathbf{I}_k$,
\begin{align}
  \|\mathbf{Q}_w-\mathbf{V}\mathbf{V}^{\top}\mathbf{Q}_w\|_F^2
  &=
  \frac{1}{s_1K_1}\|\mathbf{Q}_{\text{HF}}
  -\mathbf{V}\mathbf{V}^{\top}\mathbf{Q}_{\text{HF}}\|_F^2 +
  \frac{w}{s_2K_2}\|\mathbf{Q}_{\text{LF}}
  -\mathbf{V}\mathbf{V}^{\top}\mathbf{Q}_{\text{LF}}\|_F^2.
  \label{eq:weighted_pod_objective}
\end{align}
Consequently, the leading $k$ left singular vectors of $\mathbf{Q}_w$
minimize the weighted projection-error objective in
Eq.~\eqref{eq:weighted_pod_objective}, following
Eq.~\eqref{eq:pod_svd}. Setting $w=1$ gives equal weight to the HF
and LF mean squared projection errors, while $w=0$ recovers HF POD.
Typically, we set the weight values to be $0\leq w \leq 1$ in order
to reduce the relative contribution of the LF mean squared projection
error since we know that LF data are less accurate than HF data. The
weight $w$ is selected using the validation procedure described
in~\Cref{sec:implementation}. %

\subsection{Implementation details}
\label{sec:implementation}
We describe here the implementation choices made for the MF-OpInf
approach. First, in this work, LF snapshots are interpolated onto the
HF spatial grid, followed by any lifting, centering, and projection
used to construct the ROM. In general, we can pick any common
reference spatial grid to interpolate all the fidelity levels onto,
and the reference grid does not need to be the HF spatial grid.
Second, data from each fidelity level are projected to the same
reduced dimensional space as described in \Cref{sec:reduced}. The
reduced dimension $k$ can be chosen to retain a prescribed fraction
of the cumulative POD energy or selected using validation data, and
in this work we select it through a grid search based on validation
data. Third, for each candidate $\boldsymbol{\alpha}$ and regularizer
$\mathbf{\Gamma}$, we solve Eq.~\eqref{eq:mfopinf_operator} for the
reduced operators $\mathbf{O}$. Candidates that fail the condition
that $\mathbf{G}_{\text{MF}}+\mathbf{\Gamma}^{\top}\mathbf{\Gamma}$
is positive definite are rejected.

We select regularization parameters and, when applicable,
$\boldsymbol{\alpha}$ and the w-POD weight $w$ using a HF validation
interval immediately following the HF training horizon at the
parameters in $P_1$. Validation snapshots are excluded from operator
fitting, derivative estimation, and construction of the snapshot mean
and basis. For each candidate model, the ROM is integrated from the
initial state at the start of the ROM interval, and its reconstructed
state is compared with the HF validation snapshots. Regularization is
selected by minimizing the mean validation error across trajectories
for each fixed $(w,\boldsymbol{\alpha})$, after which the pair with
the lowest error is selected. We also ensure that candidates complete
integration through the prescribed prediction horizon with finite
states and satisfy a state growth bound, specified as a multiple of
the largest reduced HF training state norm as prescribed in
\cite{mcquarrie2021data}. This check is applied for both the HF OpInf
ROM and the MF-OpInf ROM, and it requires no HF reference data beyond
the validation interval. Note that checking for stability in this
manner is a heuristic approach and does not guarantee stability;
thus, stability-promoting constraints with
OpInf~\cite{goyal2025guaranteed,gkimisis2025representation,errico2026gasnitrom}
can be a complementary piece to the proposed MF-OpInf method.
Unseen parameter trajectories are reserved for test-time evaluation
only. Algorithm~\ref{alg:mf_opinf} summarizes the procedure for w-POD
for a given reduced dimension $k$. For HF POD, the basis is
constructed once and the search over $w$ is not necessary.
\begin{algorithm}[!htb]
  \caption{MF-OpInf with validation-based hyperparameter selection}
  \label{alg:mf_opinf}
  \begin{algorithmic}[1]
    \Require Training trajectories on $P_1\subseteq\cdots\subseteq
    P_L$, HF validation trajectories on $P_1$, reduced dimension $k$,
    and candidate values of $w$, $\boldsymbol{\alpha}$, and the
    regularization parameters.
    \For{each candidate $w$}
    \State Construct the basis $\mathbf{V}$ via w-POD (using
      Eqs.~\eqref{eq:weighted_snapshot_matrix} and
    \eqref{eq:weighted_pod_objective}).
    \State Project each trajectory, estimate time derivatives, and
    assemble $\mathbf{D}^{(l)}_{P_l}$ and $\mathbf{R}^{(l)}_{P_l}$
    using Eq.~\eqref{eq:D_and_R} for every fidelity level $l=1,\dots,L$.
    \For{each candidate $\boldsymbol{\alpha}$}
    \State Assemble $\mathbf{G}_{\text{MF}}$ using
    Eq.~\eqref{eq:mf_opinf_multi_A} and $\mathbf{F}_{\text{MF}}$
    using Eq.~\eqref{eq:mf_opinf_multi_B}.
    \For{each candidate set of regularization parameters}
    \State Form $\mathbf{\Gamma}$ and solve
    Eq.~\eqref{eq:mfopinf_operator} to get $\mathbf{O}$.
    \Comment{Reject inadmissible systems}
    \State Integrate the ROM for each $p\in P_1$ and apply the state
    growth checks. \Comment{Check on HF training data}
    \State Compute the mean HF validation error. \Comment{Assign
    rejected candidates large error}
    \EndFor
    \EndFor
    \EndFor
    \State Select hyperparameters $(w, \boldsymbol{\alpha})$ and
    regularization parameters with the lowest validation error and
    obtain corresponding basis $\mathbf{V}$ and operators $\mathbf{O}$.
    \State \Return Basis $\mathbf{V}$ and operators $\mathbf{O}$.
  \end{algorithmic}
\end{algorithm}

\section{Numerical experiments}
\label{sec:experiments}
We assess the proposed MF-OpInf method on a nonlinear
reaction-diffusion equation with a parametrized sinusoidal source
term and on flow past a cylinder with a parametrized pulse in the
inflow velocity. In both problems, one LF model provides additional
trajectories and snapshots beyond the HF training time horizon. We
compare the resulting ROMs with standard HF OpInf under a common
budget for generating training data. We first describe the allocation
of this budget, the error metrics, and the selection of model
hyperparameters followed by the numerical experiments.

\subsection{Experimental setup}
\label{sec:experiment_setup}
The HF-only OpInf baseline uses centered POD of its HF training
snapshots. Unless specified otherwise, MF-OpInf uses w-POD with unit
HF snapshot weight and LF snapshot weight as determined by the
validation procedure. The effect of the choice of the reduced basis
is shown in~\Cref{sec:basis_comp} along with the singular value decay
plots for the two numerical experiments. The search for
regularization parameters is over the range $10^{-10}$ to $10^5$ for
each parameter, $\alpha \in [0, 1]$, and $w \in [0, 1]$. Time
derivatives are estimated using fourth-order finite differences.

We consider two fidelity levels and express the cost of generating
training trajectories in equivalent HF simulation time to provide a
fair comparison. Let $c_{\text{HF}}$ and $c_{\text{LF}}$ denote the
simulation costs per unit time, and define
$C_R=c_{\text{HF}}/c_{\text{LF}}$. For $s_{\text{HF}}$ HF
trajectories simulated over a duration of $T_{\text{HF}}$ and
$s_{\text{LF}}$ LF trajectories simulated over a duration of
$T_{\text{LF}}$, the equivalent cost is
\begin{equation}
  \label{eq:equivalent_cost}
  C_{\text{eq}}=s_{\text{HF}}T_{\text{HF}}+\frac{s_{\text{LF}}T_{\text{LF}}}{C_R}.
\end{equation}
This cost model assumes that the cost scales approximately linearly
with simulated time for each model.
An HF-only baseline with $s_1$ trajectories of duration $T_1$ has
budget $C_1=s_1T_1$ and we select MF training data such that
$C_{\text{eq}}\leq C_1$. The budget accounts for simulations used to
construct the basis and fit the operators; it excludes HF validation
data, test simulations, and ROM construction and
hyperparameter-search costs. Validation data is not used for basis
construction or operator fitting, and is reported
in~\Cref{tab:heat_allocations,tab:cylinder_allocations}.

\begin{figure}[!htb]
  \centering
  \includegraphics[width=\linewidth]{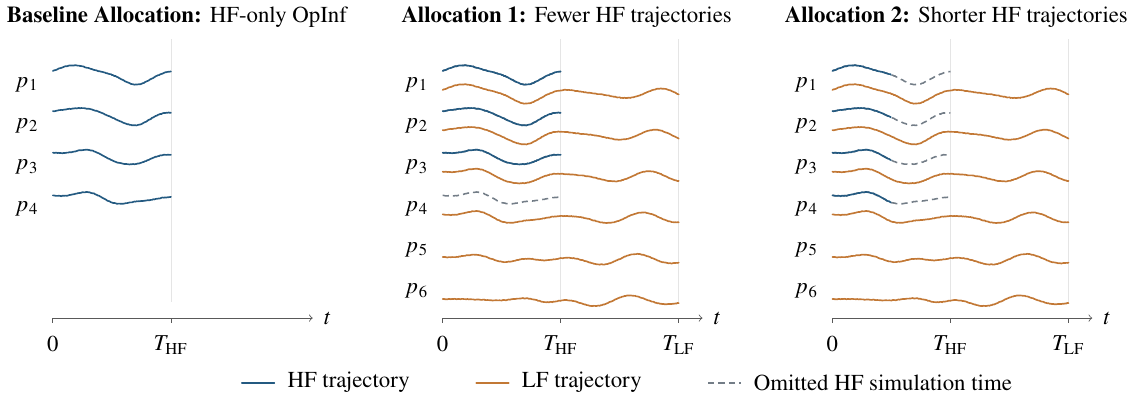}
  \caption{Comparison of the two allocation schemes employed in this
    work. Omitted HF simulation time refers to data the baseline
    HF-only OpInf is trained with, but the MF-OpInf is not. In
    Allocation 1, we evaluate the HF model for fewer trajectories as
    compared to the dataset used for baseline HF OpInf. In Allocation
    2, we evaluate the HF model for the same number of trajectories but
  shorter in time.}
  \label{fig:allocation}
\end{figure}

\paragraph{Budget allocations.} We examine two allocations of the
total computational budget. In \textit{Allocation 1}, we evaluate the
HF model at fewer parameter values while maintaining the HF
simulation duration $T_1$. The remaining budget is used to generate
LF trajectories at the shared parameters (due to the nested sampling
assumption in~\Cref{sec:MFMC_setup}) and at additional parameter
values. In \textit{Allocation 2}, we retain all $s_1$ HF parameter
values but simulate each HF trajectory for a shorter duration. The
resulting savings again fund LF trajectories. In both cases, the LF
simulations extend beyond the HF training time horizon. These
allocations are prescribed choices intended to compare parameter
coverage and temporal coverage, rather than solutions of an
allocation optimization problem. We visually depict the difference
between Allocations 1 and 2 in~\Cref{fig:allocation}. Crucially, all
ROM comparisons in the following experiments use the same training
simulation budget $C_\text{eq}$, or less, and are run on the same machine.

\paragraph{Error metric.}
\label{sec:experiment_errors}
Let $\mathbf{q}_{\text{HF}}(t;p)$ denote the HF state for a parameter
$p$, and let $\widetilde{\mathbf{q}}(t;p)$ denote the reconstructed
ROM solution. We use the relative $L^2$ error at each timestep as the
metric, given by
\begin{equation}
  \label{eq:experiment_instant_error}
  \frac{\|\mathbf{q}_{\text{HF}}(t;p)-\widetilde{\mathbf{q}}(t;p)\|_{L^2}}{\|\mathbf{q}_{\text{HF}}(t;p)\|_{L^2}}.
\end{equation}
Time-history figures show the instantaneous error in
Eq.~\eqref{eq:experiment_instant_error}. For an interval $I$, we use
the time integrated relative $L^2$ error
\begin{equation}
  \label{eq:experiment_time_error}
  \left(\frac{\int_I\|\mathbf{q}_{\text{HF}}(t;p)-\widetilde{\mathbf{q}}(t;p)\|_{L^2}^2\,\text{d}t}{\int_I\|\mathbf{q}_{\text{HF}}(t;p)\|_{L^2}^2\,\text{d}t}\right)^{1/2}.
\end{equation}
When considering error over several parameters, we
compute~\eqref{eq:experiment_time_error} for each trajectory, and
report means over parameters in tables and medians and percentile
bands in figures.

\subsection{Nonlinear reaction-diffusion equation with cubic reaction
term and sinusoidal forcing}
\label{sec:cubic_heat}
We consider the forced cubic reaction-diffusion equation, as studied
in~\cite{mcquarrie2025bayesian}. Nonlinear reaction-diffusion
equations arise in a variety of applications, including chemical
kinetics and combustion processes~\cite{xin2000front}. In this work,
we consider an example of such a process with the following model form:
\begin{equation}
  \label{eq:heat_pde}
  \frac{\partial q}{\partial t}=\mu\frac{\partial^2q}{\partial
  x^2}-q^3+u(x,t;a,b),\qquad x\in(0,1),\quad t\in(0,2],
\end{equation}
with diffusion coefficient $\mu=0.005$, Dirichlet boundary conditions
on the temperature $q(0,t)=0$ and $q(1,t)=1$, and initial condition
\begin{equation}
  q(x,0)=x(1-x)\left(6(1-x)^2e^{-x}-10e^x\sin(x/6)\right)+x.
\end{equation}
The source term is
\begin{equation}
  \label{eq:heat_source}
  u(x,t;a,b)=\frac{a\sin(2\pi t)}{1+100(x-1/4)^2}+\frac{b\sin(4\pi
  t)}{1+100(x-3/4)^2},
\end{equation}
where $p=(a,b)\in[-2,2]^2$. The two spatial source profiles are
centered at $x=1/4$ and $x=3/4$ and oscillate at different
frequencies. Varying their amplitudes produces multiple trajectories
with a common initial condition.

\begin{figure}[!htb]
  \centering
  \includegraphics[width=\linewidth]{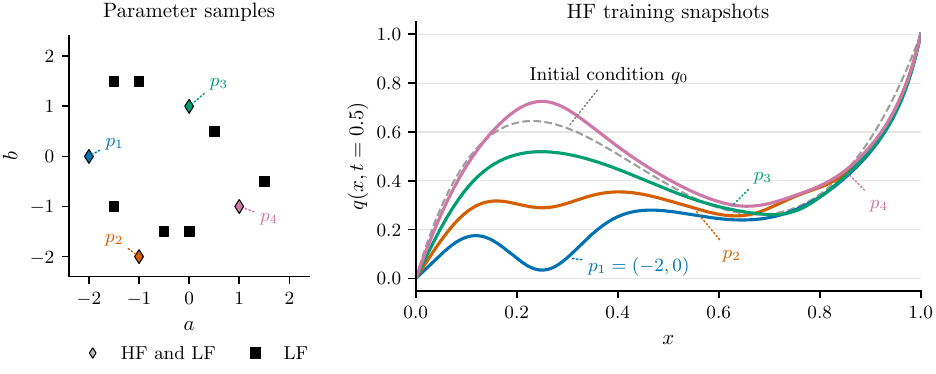}
  \caption{Illustrative diagram of parameter samples and HF states
    for the nonlinear reaction-diffusion problem. Left: the four
    baseline HF parameters, LF parameters from the two allocations, and
    additional LF candidates. Right: the common initial condition and
    HF profiles at $t=0.5~\mathrm{s}$, with colors corresponding to the
  HF parameters.}
  \label{fig:heat_parameters}
\end{figure}

The HF and LF models use uniform spatial grids with 401 and 101
points, respectively. The HF relative and absolute solver tolerances
are $10^{-10}$ and $10^{-12}$, while the LF tolerances are $10^{-6}$
and $10^{-8}$. Snapshots are stored at uniform intervals $\Delta
t=1/800$. LF states are interpolated linearly onto the HF grid before
lifting and projection. This leads to an HF-to-LF simulation cost
ratio $C_R=12$ in Eq.~\eqref{eq:equivalent_cost} based on wall-clock
times. The HF-only baseline uses four trajectories with parameter
values $P_1=\{(-2,0),(-1,-2),(0,1),(1,-1)\}$ over $t\in[0,1]$ s.
Allocation 1 uses the first three HF parameter values over the same
interval and six LF trajectories over $t\in[0,2]$ s. Allocation 2
uses all four HF parameter values over $t\in[0,0.7]$ s and seven LF
trajectories over $t\in[0,2]$ s. The LF parameter set contains the HF
parameter set due to the nested samples setup. In both budget
allocations, the LF snapshots extend to $T_\mathrm{LF} = 2$ s, which
is beyond the HF training time horizon $T_\mathrm{HF} = 1$ s. The
resulting equivalent training data costs are summarized in
\Cref{tab:heat_allocations}. Allocation 1 uses the full budget of
four equivalent HF time units. Allocation 2 uses $3.967$ units,
leaving $0.033$ units unspent because another complete LF trajectory
would exceed the budget.
The validation interval has a duration $0.2$ s after the training time horizon.
For test data, we generate HF trajectories through $t=2$ s at the
four parameters in $P_1$ and at 20 additional test parameter values
sampled uniformly from $[-2,2]^2$. The parameter coverage and
representative spatial profiles are illustrated in \Cref{fig:heat_parameters}.
Prediction after $t=1$ s extends beyond the HF training data, while
LF training data remain available through $t=2$ s.
\begin{table}[!htb]
  \centering
  \caption{Training data allocation for the nonlinear
    reaction-diffusion equation. All methods have a budget of four
  equivalent HF simulation time units.}
  \label{tab:heat_allocations}
  \begin{tabular}{lcccccc}
    \hline
    Method & $s_{\text{HF}}$ & $T_{\text{HF}}$ (s) & $s_{\text{LF}}$
    & $T_{\text{LF}}$ (s) & $C_{\text{eq}}$ & Validation Cost \\
    \hline
    HF OpInf & 4 & 1 & 0 & -- & 4.000 & 0.8 \\
    MF-OpInf (Alloc 1) & 3 & 1 & 6 & 2 & 4.000 & 0.6 \\
    MF-OpInf (Alloc 2) & 4 & 0.7 & 7 & 2 & 3.967 & 0.8 \\
    \hline
  \end{tabular}
\end{table}

\paragraph{Reduced model and training.}
To expose quadratic dynamics and re-express the cubic nonlinearity in
Eq.~\eqref{eq:heat_pde}, we introduce an auxiliary variable through
the lifting
procedure~\cite{qian2020lift,kramer2019nonlinear,benner2015two}. We
lift the spatially discretized state to
$[\mathbf{q}^{\top},(\mathbf{q}\odot\mathbf{q})^{\top}]^{\top}$,
where $\odot$ denotes the Hadamard (elementwise) product, yielding
802 lifted state variables on the HF grid. Centering and projecting
the lifted snapshots yields the reduced model
\begin{equation}
  \label{eq:heat_rom}
  \frac{\text{d}}{\text{d}t}\mathbf{z}(t)=\widehat{\mathbf{c}}+\widehat{\mathbf{A}}\mathbf{z}(t)+\widehat{\mathbf{H}}[\mathbf{z}(t)\otimes\mathbf{z}(t)]+\widehat{\mathbf{B}}\mathbf{u}(t)+\widehat{\mathbf{N}}[\mathbf{u}(t)\otimes\mathbf{z}(t)].
\end{equation}
The constant term accounts for centering, the state-input term
$\widehat{\mathbf{N}} \in \mathbb{R}^{k\times km}$ arises from the
lifting, and $\mathbf{u}(t;a,b) = (a\mathrm{sin}(2\pi
t),b\mathrm{sin}(4\pi t))^\top \in \mathbb{R}^2$ are the
time-dependent components of the source term $u$. The additional
term, $\widehat{\mathbf{N}}$, introduces bilinear interactions
between the state and input, and showcases the flexibility of OpInf
to incorporate varying model forms. A single set of reduced operators
is learned across all training trajectories since parameters only
enter the governing equation through $\mathbf{u}(t;a,b)$.
We regularize three groups of operators:
$(\widehat{\mathbf{c}},\widehat{\mathbf{A}})$,
$(\widehat{\mathbf{B}},\widehat{\mathbf{N}})$, and
$\widehat{\mathbf{H}}$, where each is regularized separately. The ROM
is initialized by quadratic lifting and projection of the HF initial
state at $t=0$, and integrated using adaptive BDF with relative and
absolute tolerances $10^{-8}$ and $10^{-10}$, respectively.

\begin{figure}[!htb]
  \centering
  \begin{subfigure}[t]{0.49\textwidth}
    \centering
    \includegraphics[width=\linewidth]{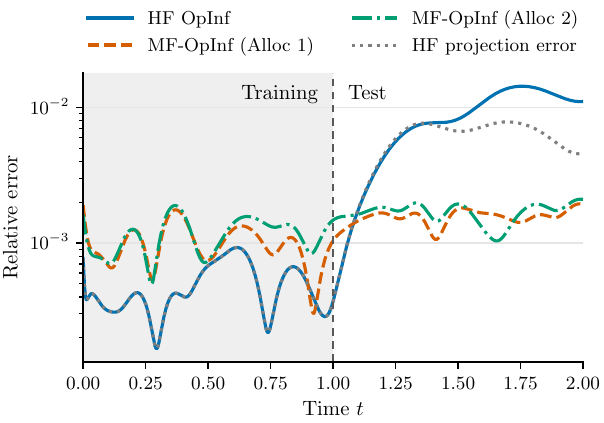}
    \caption{Training parameter: $p=(-1.0,\,-2.0)$}
    \label{fig:heat_time_errors_left}
  \end{subfigure}
  \hfill
  \begin{subfigure}[t]{0.49\textwidth}
    \centering
    \includegraphics[width=\linewidth]{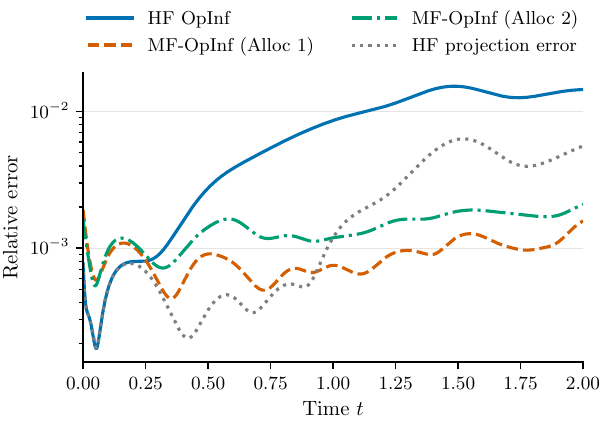}
    \caption{Test parameter: $p=(-0.681,\,1.154)$}
    \label{fig:heat_time_errors_right}
  \end{subfigure}
  \caption{Relative prediction errors for $k=21$ at equal training
    simulation cost. Instantaneous relative errors for a training and a
    test parameter, including the HF POD reconstruction error. Gray
    shading marks the HF training interval $t \in [0,1]~\mathrm{s}$. LF
    training extends through $t=2~\mathrm{s}$. HF projection error
    refers to the HF-only POD basis used by the HF OpInf baseline.
  MF-OpInf uses the w-POD basis.}
  \label{fig:heat_time_errors}
\end{figure}

\paragraph{Temperature field prediction.}
We compare the HF-only OpInf with the two MF-OpInf approaches, where
the reduced dimension, regularization parameters, and hyperparameter
values were selected via the validation procedure described
in~\Cref{sec:implementation}. This led to $\alpha=0.1$ for Allocation
1 and $\alpha=0.25$ for Allocation 2. Both had weight $w=0.5$ for the
w-POD basis. The reduced dimension was selected to be $k=21$ for all
three approaches. At representative training and test parameters, we
show the relative error of the MF-OpInf method in comparison with HF
OpInf and the projection error of the HF POD basis that HF OpInf used
in~\Cref{fig:heat_time_errors_left,fig:heat_time_errors_right},
respectively. In general, MF-OpInf trades some accuracy during the
training interval at training parameters for improved predictions at
later times for both training and testing parameter values.
\Cref{tab:heat_errors} reports the mean relative errors across the
four training parameters and the 20 test parameters.
At the test parameters, MF-OpInf reduces the mean error over the test
interval $t\in[1,2]$ s by approximately $87\%$, which captures the
reduction in prediction error during temporal and parametric extrapolation.
The HF-only OpInf has the smaller error over $t\in[0,1]$ s at the
training parameters, but its error increases beyond the HF training
horizon, consistent with possible overfitting to the training regime.
Thus, the principal benefit of the MF data in this comparison is
improved prediction at test parameters and later times, with the same
training simulation cost.
\begin{table}[!htb]
  \centering
  \caption{Mean relative $L^2$ error across parameters in percent for
    $k=21$ at equal training simulation cost. The error is computed
    over each time interval for each trajectory before averaging across
    parameters. MF-OpInf uses Allocation 1, w-POD, and the
    validation-selected coefficient $\alpha=0.1$. Bolded numbers
    represent the lowest error for that parameter set and time interval
  (lower is better).}
  \label{tab:heat_errors}
  \begin{tabular}{llccc}
    \hline
    Model & Parameter set & $t \in [0,1]$ s & $t \in [1,2]$ s & $t
    \in [0,2]$ s \\
    \hline
    HF OpInf & Training & \textbf{0.044} & 1.168 & 0.788 \\
    MF-OpInf (Alloc 1) & Training & 0.091 & \textbf{0.125} & \textbf{0.109} \\
    HF OpInf & Test & 0.261 & 0.972 & 0.680 \\
    MF-OpInf (Alloc 1) & Test & \textbf{0.100} & \textbf{0.130} &
    \textbf{0.115} \\
    \hline
  \end{tabular}
\end{table}

\begin{figure}[!htb]
  \centering
  \includegraphics[width=\textwidth]{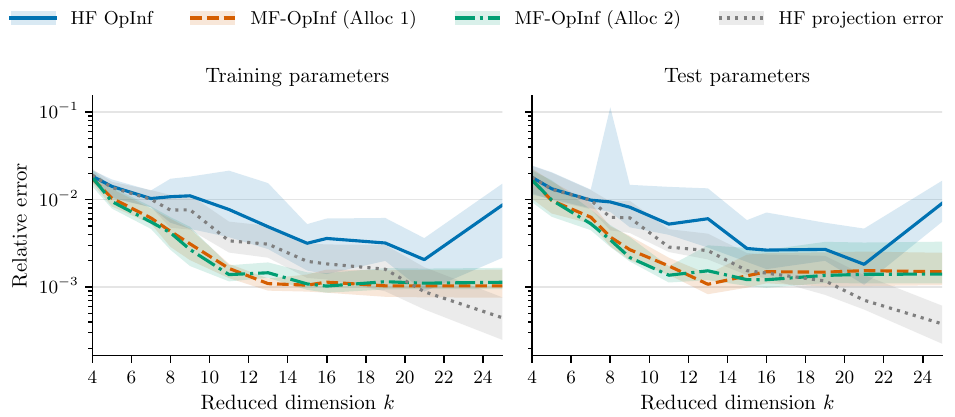}
  \caption{Relative $L^2$-in-time error over the entire interval $t
    \in [0,2]~\mathrm{s}$ versus reduced dimension at equivalent
    training simulation budget. Lines denote medians and shaded bands
    span the 10th to 90th percentiles across the four training
    parameters (left) and 20 test parameters (right). HF projection
    error refers to the HF-only POD basis used by the HF OpInf
  baseline. MF-OpInf uses the w-POD basis.}
  \label{fig:heat_dimension}
\end{figure}

\paragraph{Effect of reduced dimension.}
We next compare both budget allocations over different reduced
dimension sizes. Regularization and $\alpha$ are reselected through
validation error at each dimension. \Cref{fig:heat_dimension} shows
the median error over $t \in [0,2]$ s and the 10th to 90th percentile
range across each evaluation parameter set. Both MF allocations
improve prediction over the HF-only baseline at intermediate and
larger dimensions, with errors approaching a plateau. MF-OpInf also
leads to a more robust fit compared to HF-only OpInf. The HF POD
reconstruction error continues to decrease as the dimension
increases, but the HF-only prediction error does not decrease. This
is expected since a larger reduced-dimensional space alone does not
ensure more accurate inferred dynamics, which also depends on the
available training snapshots.

\subsection{Flow past a cylinder with a parametrized inlet pulse}
\label{sec:cylinder}
We consider two-dimensional incompressible flow past a cylinder. A
version of this problem has been commonly used in the ROM
literature~\cite{schafer1996benchmark,
gunzburger2019evolve,mohebujjaman2019physically,strazzullo2022consistency,wells2017evolve,moore2026learning}.
However, our setup is adapted directly from publicly available code
developed by Moore et al.~\cite{moore2026learning}, where the
cylinder is a sizable proportion of the inlet width. This is
reflective of the geometry defined in~\cite{schafer1996benchmark}. In
this version of the problem, the flow starts at rest and is ramped up
until $t=4$ s. Then, snapshots from a developing wake from $t \in
[4,6]$ s are used to predict the subsequent flow from $t\in [10, 20]$
s. This is in contrast to training and prediction within the same
statistically steady regime. As is noted in~\cite{moore2026learning},
this is not a traditional setting for ROMs on flow-past-a-cylinder
problems and leads to more difficult extrapolation in time for the
inferred operators. In this work, we follow this setup and in
addition introduce a parametrized pulse at the inlet before the
training time horizon, which alters the initial flow field at $t=4$ s
and allows us to generate data consisting of multiple trajectories.
However, because of the nature of the pulse, this problem primarily
tests temporal extrapolation from training on a developing flow to
predicting the established wake. We examine whether additional LF
trajectories improve prediction (in time and for new parameters)
beyond a short HF training interval at a fixed simulation budget.

The governing equations for this problem are the 2D incompressible
Navier-Stokes equations,
\begin{equation}
  \label{eq:cylinder_pde}
  \begin{aligned}
    \frac{\partial\mathbf{u}}{\partial
    t}+(\mathbf{u}\cdot\nabla)\mathbf{u}-\nu\Delta\mathbf{u}+\frac{1}{\rho}\nabla
    P&=\frac{1}{\rho}\mathbf{f},\\
    \nabla\cdot\mathbf{u}&=0,
  \end{aligned}
  \qquad (x,y)\in\Omega,\quad t\in[0,20],
\end{equation}
where $\mathbf{u}=(u_x,u_y)^{\top}$ is the velocity, $P$ is the
pressure, $\rho=1$ is the constant fluid density, and
$\nu=\mu/\rho=10^{-3}$ is the kinematic viscosity. Previously
$\mathbf{u}$ was defined as the input terms and parameters in the
ROM. In this section, we will explicitly state when we are referring
to the physical quantity of velocity in Eq.~\eqref{eq:cylinder_pde}
for expositional clarity. In this application,
$\mathbf{f}=\mathbf{0}$, and the flow is driven by the prescribed
inlet velocity. The domain $\Omega$ is a channel of length $2.2$ and
height $h=0.41$, excluding a cylinder shaped region of radius
$R=0.05$ centered at $(0.2,0.2)$. We impose no-slip conditions on the
channel walls and cylinder, and zero pressure at the outlet. The
horizontal inlet velocity is a parabolic profile given by,
\begin{equation}
  \label{eq:cylinder_inlet}
  u_x(t,y;a)=
  \begin{cases}
    \dfrac{6y(h-y)}{h^2}\sin(\pi t/8), & 0\leq t<2, 3\leq t<4,\\[6pt]
    \dfrac{6y(h-y)}{h^2}\left[\sin(\pi
    t/8)+a\sin^2\bigl(\pi(t-2)\bigr)\right], & 2\leq t<3,\\[6pt]
    \dfrac{6y(h-y)}{h^2}, & t\geq4,
  \end{cases}
\end{equation}
and the vertical inlet velocity $u_y$ is set to zero. The parameter
$a \in [-0.225, 0.225]$ controls the pulse and is varied to generate
multiple trajectories. We choose this profile so that the pulse
smoothly increases in magnitude and returns to zero before the
training interval begins at $t=4$ s. The pulse acts only during
$2<t<3$ and reaches its maximum magnitude at $t=2.5$ s, where its
value is $a$. For $t\geq4$ s, all trajectories have the same steady
inlet profile. The Reynolds number for this flow based on the
cylinder diameter is $\mathrm{Re}=100$.

The code is implemented in FEniCSx. %
To create an LF version of this simulation, we adjust the mesh
resolution and solver timestep, as summarized in
\Cref{tab:cylinder_discretization}. Training data snapshots are
stored at common intervals $\Delta t=0.005$ beginning at $t=4$ s. LF
velocity snapshots are interpolated onto the HF velocity space before
centering and projection.
The resulting cost ratio was $C_R=20$.
The HF-only OpInf uses four parameters $P_1=\{-0.2,0,0.1,0.2\}$,
which adjust the pulse, with HF snapshots on training time horizon $t
\in [4,6]$ s. Allocation 1 retains the three HF parameters
$\{0,0.1,0.2\}$ over the same interval and uses additional LF
trajectories. The LF snapshots are obtained on the training time
horizon $t \in [4,8]$ s, which extends beyond the HF training
horizon. This results in 401 HF snapshots and 801 LF snapshots per trajectory.
\Cref{tab:cylinder_allocations} reports the resulting cost of the
training data. All three configurations use 24 equivalent HF
simulation time units. This cost includes the flow startup interval
$t\in [0,4]$ s, even though that data is not used for operator
fitting or basis construction.
For validation data, the HF OpInf baseline and MF-OpInf Allocation 1
use HF snapshots on $t \in [6,10]$ s at their respective HF training
parameters. Allocation 2 uses $t \in [5,9]$ s because its HF training
ends at $t=5$ s. The validation window $t \in [6, 10]$ s lies in the
still-developing flow and is deliberately withheld from training for
all methods, so that the established shedding regime in $t \in [10,
20]$ s is reached only by extrapolation. For test data, HF
trajectories extend through $t=20$ s at the four training parameters
in $P_1$ and at 12 additional test parameters.
\begin{table}[!htb]
  \centering
  \caption{Spatial and temporal discretizations for the cylinder
    problem. Both fidelities use a prescribed far-field element size
  $h/4=0.1025$ and snapshot spacing $\Delta t_{\mathrm{snap}}=0.005$.}
  \label{tab:cylinder_discretization}
  \begin{tabular}{lcccc}
    \hline
    Fidelity & Near-cylinder target element size & Solver timestep &
    Velocity DOFs & Pressure DOFs \\
    \hline
    HF & $R/3$ & $0.0005$ & 16,048 & 2,060 \\
    LF & $R/1.5$ & $0.005$ & 7,440 & 972 \\
    \hline
  \end{tabular}
\end{table}
\begin{table}[!htb]
  \centering
  \caption{Training data allocation for the cylinder problem.
    Simulation durations include startup from rest at $t=0~\mathrm{s}$,
  while ROM training begins at $t=4~\mathrm{s}$.}
  \label{tab:cylinder_allocations}
  \begin{tabular}{lcccccc}
    \hline
    Method & $s_{\text{HF}}$ & $T_{\text{HF}}$ (s) & $s_{\text{LF}}$
    & $T_{\text{LF}}$ (s) & $C_{\text{eq}}$ & Validation Cost \\
    \hline
    HF OpInf & 4 & 6 & 0 & -- & 24 & 16 \\
    MF-OpInf (Alloc 1) & 3 & 6 & 15 & 8 & 24 & 12 \\
    MF-OpInf (Alloc 2) & 4 & 5 & 8 & 10 & 24 & 16 \\
    \hline
  \end{tabular}
\end{table}

\paragraph{Reduced model and training.}
We approximate the HF velocity by $\mathbf{u}(t;a)\approx
\overline{\mathbf{u}}+\mathbf{V}\mathbf{z}(t;a)$,
where $\overline{\mathbf{u}}$ is the mean velocity, and infer a quadratic ROM,
\begin{equation}
  \label{eq:cylinder_rom}
  \frac{\text{d}}{\text{d}t}\mathbf{z}(t)=\widehat{\mathbf{c}}+\widehat{\mathbf{A}}\mathbf{z}(t)+\widehat{\mathbf{H}}[\mathbf{z}(t)\otimes\mathbf{z}(t)].
\end{equation}
The pulse changes the HF state at $t=4$ s, but has ended before the
ROM interval begins. Since all trajectories have the same steady
inlet, the ROM has no time-dependent input term. We infer one set of
operators across the training trajectories using snapshots of the
velocity field. We regularize two groups of operators:
$(\widehat{\mathbf{c}},\widehat{\mathbf{A}})$ and
$\widehat{\mathbf{H}}$. The ROM is initialized by projecting the
corresponding HF state at $t=4$ s and integrated with adaptive RK45
using relative and absolute tolerances $10^{-6}$ and $10^{-8}$.

\begin{figure}[!htb]
  \centering
  \includegraphics[width=\linewidth]{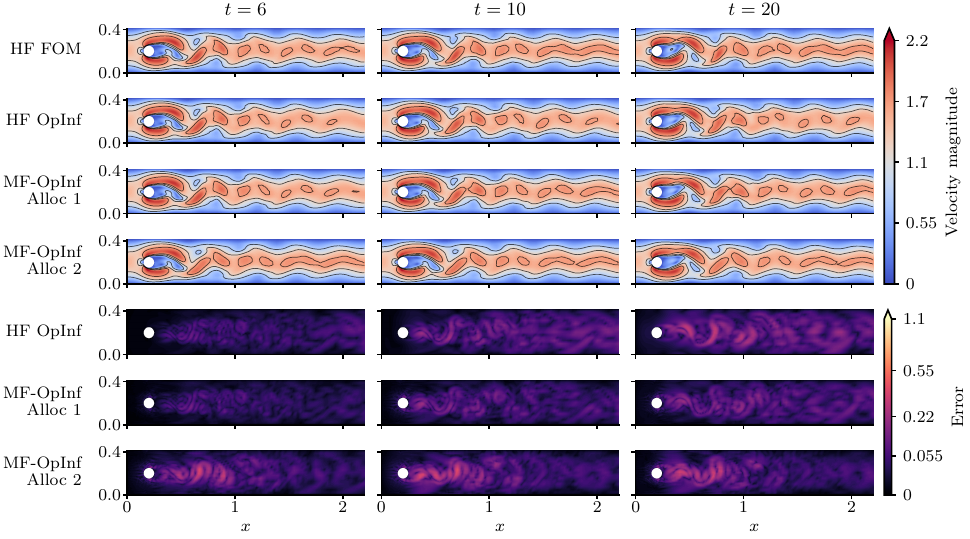}
  \caption{Velocity fields at the test parameter $a=0.15$ at equal
    training simulation cost. Columns correspond to $t=6,10,20$ s. The
    top four rows show the velocity magnitude for the HF reference FOM
    and the three ROMs, with black contours indicating constant
    velocity magnitudes. The bottom three rows show the magnitude of
  the velocity error relative to the HF reference.}
  \label{fig:cylinder_fields}
\end{figure}

\paragraph{Velocity field predictions.}
We first compare the reconstructed velocity fields at the test
parameter $a=0.15$. As in the previous application, the
hyperparameters and ROM configurations were selected using validation
error only for both the HF and MF-OpInf ROMs. The validation
selection procedure led to the HF-only approach and MF-OpInf with
Allocation 1 using $k=21$, while Allocation 2 uses $k=16$.
Extrapolation in time was a key feature and difficulty in this
problem, and thus the Allocation 2 approach for MF-OpInf suffered,
because the training data consists of HF snapshots for a shorter
duration in time.
For both MF-OpInf approaches, we construct a shared w-POD basis from
the HF snapshots and interpolated LF snapshots, with HF and LF
weights of 1 and $w=0.1$, respectively. Both MF approaches selected
an $\alpha=0.001$. Although the selected coefficient is small, the
magnitude of $\alpha$ alone does not indicate the influence of the LF
correction.
\Cref{fig:cylinder_fields} shows the velocity magnitude and the
magnitude of the velocity error at $t=6,10,20$ s. To facilitate
visual comparison of the flow structures, we overlay contours of
constant velocity magnitude at levels of 0.8, 1.2, and 1.6 on the
velocity fields. All three ROMs reproduce the main wake structure. At
$t=20$ s, MF-OpInf with Allocation 1 has smaller errors in the wake
than the HF-only OpInf. MF-OpInf with Allocation 2 has larger errors
at $t=6$ s, but its later predictions still recover the main flow features.

\begin{figure}[!htb]
  \centering
  \begin{subfigure}[t]{0.49\textwidth}
    \centering
    \includegraphics[width=\linewidth]{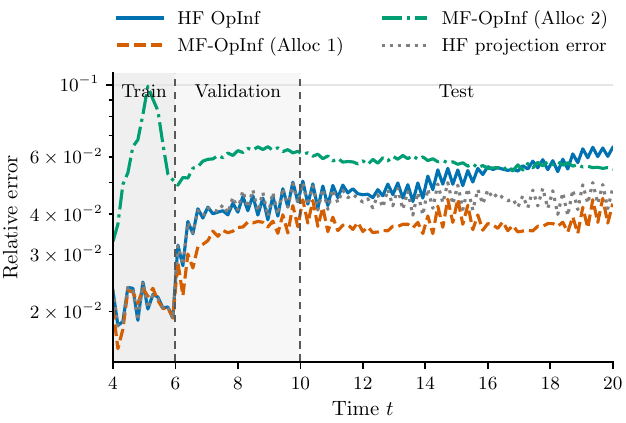}
    \caption{Training parameter: $a=0.10$}
    \label{fig:cylinder_time_errors_left}
  \end{subfigure}
  \hfill
  \begin{subfigure}[t]{0.49\textwidth}
    \centering
    \includegraphics[width=\linewidth]{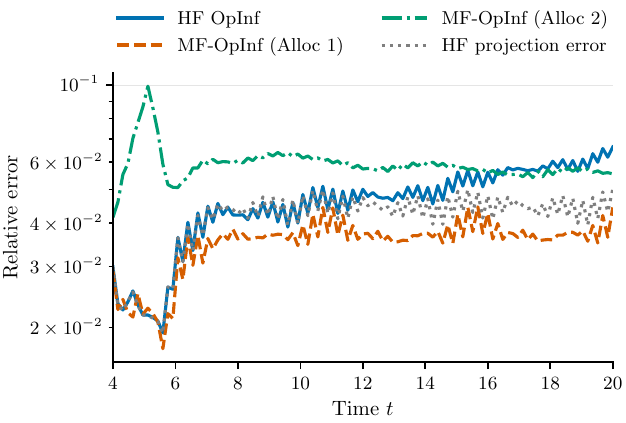}
    \caption{Test parameter: $a=0.15$}
    \label{fig:cylinder_time_errors_right}
  \end{subfigure}
  \caption{Instantaneous relative errors at equal training simulation
    cost, including the HF POD projection error. In the left subfigure,
    shading and vertical lines mark the HF training interval $t \in
    [4,6]~\mathrm{s}$ and validation interval $t \in [6,10]~\mathrm{s}$
    for the HF-only baseline and Allocation 1. Allocation 2 uses HF
    training on $t \in [4,5]~\mathrm{s}$ and validation on $t \in
    [5,9]~\mathrm{s}$. LF training extends through $t=8~\mathrm{s}$ for
    Allocation 1 and $t=10~\mathrm{s}$ for Allocation 2. HF projection
    error refers to the HF-only POD basis used by the HF OpInf
  baseline. MF-OpInf uses the w-POD basis.}
  \label{fig:cylinder_time_errors}
\end{figure}

\paragraph{Prediction accuracy over time.}
\Cref{fig:cylinder_time_errors} compares instantaneous relative
errors at the training parameter $a=0.10$ and the unseen test
parameter $a=0.15$. For both parameters, MF-OpInf with Allocation 1
maintains smaller errors than the HF-only OpInf over almost all of
the prediction interval beyond $t=6$ s. Its error remains near $4\%$
at later times, while the HF-only error increases to approximately
$6\%$ by $t=20$ s. For this problem, MF-OpInf with Allocation 2
exhibits a larger transient error near $t=5.5$ s and remains less
accurate than Allocation 1 at later times. Thus, MF-OpInf with
Allocation 1 improves temporal extrapolation at both parameters under
the same training simulation budget. In~\Cref{tab:nse_errors}, we
report the mean relative errors across the training and test
parameters. We see that the MF-OpInf with the Allocation 1 budget
strategy reduced the test regime errors on test parameters by
approximately $28\%$.
\begin{table}[!htb]
  \centering
  \caption{Mean relative spatial $L^2$ error in percent for the HF
    OpInf and MF-OpInf (Allocation 1) models. For each parameter, the
    relative error is aggregated over the indicated time interval as
    shown in~\eqref{eq:experiment_time_error}, and the table then
    averages across parameters. The models have equal training
    simulation cost. Bold indicates the lower error within each
  parameter set and interval (lower is better).}
  \label{tab:nse_errors}
  \begin{tabular}{llccc}
    \hline
    Model & Parameter set & $t\in[4,6]$ s & $t\in[6,10]$ s & $t\in[10,20]$ s \\
    \hline
    HF OpInf & Training & \textbf{2.184} & 4.047 & 5.293 \\
    MF-OpInf (Alloc 1) & Training & 3.377 & \textbf{3.698} & \textbf{3.770} \\
    HF OpInf & Test & \textbf{2.144} & 4.003 & 5.266 \\
    MF-OpInf (Alloc 1) & Test & 3.520 & \textbf{3.697} & \textbf{3.767} \\
    \hline
  \end{tabular}
\end{table}

\begin{figure}[!t]
  \centering
  \includegraphics[width=\textwidth]{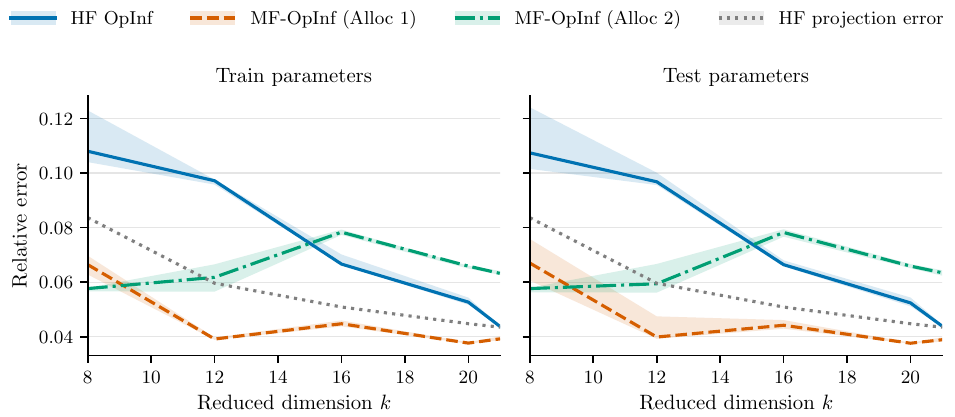}
  \caption{Relative $L^2$ error over the testing interval $t \in
    [10,20]~\mathrm{s}$ versus reduced dimension at equal training
    simulation cost. Lines denote medians and shaded bands span the
    10th to 90th percentiles across the four training parameters (left)
    and 12 test parameters (right). HF projection error refers to the
    HF-only POD basis used by the HF OpInf baseline. MF-OpInf uses the
  w-POD basis.}
  \label{fig:cylinder_dimension}
\end{figure}

\paragraph{Effect of reduced dimension.}
\Cref{fig:cylinder_dimension} shows the median relative error over $t
\in [10,20]$ s and the 10th to 90th percentile range across the four
training parameters and the 12 test parameters. For each choice of
reduced dimension, the MF-OpInf hyperparameters are selected using
validation error. MF-OpInf with Allocation 1 has the smallest median
error at most dimensions and remains near $4\%$ for $k\geq12$.
MF-OpInf with Allocation 2 improves on the HF-only OpInf at smaller
dimensions, but its error does not decrease consistently as the
dimension increases in part due to the validation selection
procedure. The HF-only error decreases through $k=21$ and increases
beyond it (not shown). As in the nonlinear reaction-diffusion
problem, increasing the reduced dimension alone does not ensure more
accurate inferred dynamics. We also observe errors below the HF
projection error, as MF-OpInf uses w-POD to incorporate LF training
data with broader parametric coverage and longer time horizons,
yielding a more accurate reduced basis than HF-only POD.

\subsection{Contributions of LF data to the basis construction and
the operator fitting}
\label{sec:basis_comp}
MF-OpInf uses LF data in two stages: to construct the reduced basis
through w-POD (Section~\ref{sec:reduced}) and to fit the reduced
operators through the MF loss (Section~\ref{sec:mf_opinf_bg}). To
isolate the contributions of these two uses of LF data, we compare
four combinations of basis construction and operator fitting at a
fixed reduced dimension $k=21$ and a fixed training simulation budget
in Table~\ref{tab:basis_comparison}. The four configurations range
from HF OpInf baseline, which uses only HF data for both stages, to
MF-OpInf, which uses LF data for both stages. The two intermediate
configurations use LF data in only one stage.

\begin{figure}[!htb]
  \centering
  \begin{subfigure}[t]{0.49\textwidth}
    \centering
    \includegraphics[width=\linewidth]{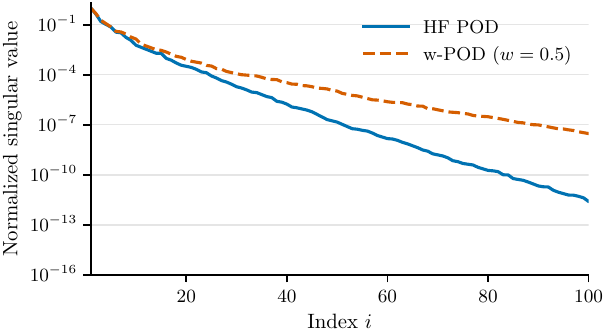}
    \caption{Reaction-diffusion equation}
    \label{fig:rd_dim}
  \end{subfigure}
  \hfill
  \begin{subfigure}[t]{0.49\textwidth}
    \centering
    \includegraphics[width=\linewidth]{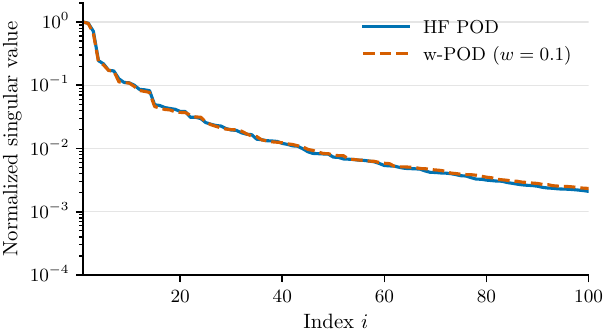}
    \caption{Flow past a cylinder}
    \label{fig:cyl_dim}
  \end{subfigure}
  \caption{Normalized singular value decay for the nonlinear
    reaction-diffusion equation and cylinder problems over the first
  100 modes of both HF POD and w-POD.}%
  \label{fig:svd}
\end{figure}

First, we look at the singular value decay for each problem
in~\Cref{fig:svd}. We note that the reaction-diffusion spectrum
decays rapidly as seen in~\Cref{fig:rd_dim}, so the reduced
dimensions we probe in this work
retain nearly all of the snapshot energy. In contrast, the cylinder
spectrum in~\Cref{fig:cyl_dim} decays more slowly over the dimensions
considered. This is to be expected given the nature of developing
flow in the flow past a cylinder problem. Given the reduced
dimension, the acceptable level of projection error is
problem-dependent and to be decided by the practitioner. In this work
we study the effect of the basis construction at a fixed $k$.

In general, for both problems, using LF data in either the basis
construction or operator fitting reduces the error relative to the HF
OpInf baseline. The only exception is that using w-POD with HF-only
operator fitting increases the error relative to the baseline for the
cylinder flow problem. For the cylinder flow problem, using MF
operator fitting with an HF POD basis reduces the error by
approximately 16\%, from 5.266\% to 4.425\%, and using LF data in
both stages reduces it by approximately 28\%, to 3.767\%.
In both problems, the lowest error is obtained when LF data are used
for both the basis construction and the operator fitting,
corresponding to the MF-OpInf configuration used throughout this work.

\begin{table}[!htbp]
  \centering
  \caption{Mean relative $L_2$ error over the test parameters and
    test interval. All models use equal training simulation budgets. MF
    configurations use Allocation 1. Bold indicates the lowest error
  for each problem.}%
  \label{tab:basis_comparison}
  \begin{tabular}{llcc}
    \hline
    & & \multicolumn{2}{c}{Relative error ($\%$)} \\
    \cline{3-4}
    Problem & Operator fitting & Basis using HF data (HF POD) & Basis
    using HF+LF data (w-POD) \\
    \hline
    \multirow{2}{*}{Reaction-diffusion}
    & HF only & 0.972 & 0.512 \\
    & MF      & 0.329 & \textbf{0.130} \\
    \hline
    \multirow{2}{*}{Flow past a cylinder}
    & HF only & 5.266 & 6.428 \\
    & MF      & 4.425 & \textbf{3.767} \\
    \hline
  \end{tabular}
\end{table}

\section{Conclusions}
\label{sec:conclusion}
This paper presents MF-OpInf, an MF formulation for OpInf ROMs, that
combines HF and LF trajectories directly in the regression objective
to learn one set of reduced operators. By providing additional
parameter coverage and extending beyond the HF training time horizon,
LF trajectories improved prediction at test parameters and later
times under a fixed training simulation budget.
We analyzed two approaches for allocating budget between HF and LF
training data trajectories. When we reallocated the cost of one HF
training trajectory towards additional LF trajectories at multiple
parameters and a longer training horizon (Allocation 1), MF-OpInf
reduced the mean relative prediction error at unseen parameters
beyond the HF horizon by approximately 87\% for the
reaction-diffusion problem and 28\% for cylinder flow. These
improvements were obtained at equal training simulation cost to the
baseline HF OpInf. We also analyzed the separate contributions of LF
data to error reduction through two stages: (i) fitting the reduced
operators through an MF loss and (ii) obtaining the reduced basis
using weighted POD. In both problems, MF-OpInf lowered the error for
either choice of basis, and the lowest error came from using
low-fidelity data for both the basis and the operator fit.

We note that the improvements depended on the training data
allocation and ROM hyperparameters. Shortening the HF trajectories
(Allocation 2) produced less consistent gains, particularly for
cylinder flow, and validation did not always select the
configurations with the lowest test error. Future work will therefore
investigate more reliable selection of the MF coefficients,
regularization parameters, and basis weights, together with
systematic allocation of the simulation budget across trajectory
counts, durations, and fidelity levels.
Further directions include adapting MF
POD~\cite{aretz2026multifidelity} to construct the reduced basis and
combining MF-OpInf with other stability-promoting techniques.

\section*{Acknowledgments}
This work has been supported in part by AFOSR grant FA9550-24-1-0327
under the Multidisciplinary University Research Initiative (MURI) and
DARPA TRS award HR00112590114. Generative artificial intelligence
(AI) tools were used to assist in refining the TikZ code for Figs.
1-3. The figures were originally designed by the authors.

\bibliography{references}

\end{document}